%% file: kdd.tex
\documentclass[sigconf]{acmart}
\usepackage{amsthm}
\usepackage{booktabs}
\usepackage{caption}
\usepackage{subcaption}
\usepackage{graphicx}
\usepackage[linesnumbered,ruled]{algorithm2e}
\usepackage{setspace}

\theoremstyle{definition}
\newtheorem{definition}{Definition}[section]
\input{math_commands.tex}

\AtBeginDocument{%
  \providecommand\BibTeX{{%
    \normalfont B\kern-0.5em{\scshape i\kern-0.25em b}\kern-0.8em\TeX}}}

\copyrightyear{2021} 
\acmYear{2021} 
\setcopyright{rightsretained} 
\acmConference[KDD '21]{Proceedings of the 27th ACM SIGKDD Conference on Knowledge Discovery and Data Mining}{August 14--18, 2021}{Virtual Event, Singapore}
\acmBooktitle{Proceedings of the 27th ACM SIGKDD Conference on Knowledge Discovery and Data Mining (KDD '21), August 14--18, 2021, Virtual Event, Singapore}
\acmDOI{10.1145/3447548.3467373}
\acmISBN{978-1-4503-8332-5/21/08}

\begin{document}
\fancyhead{}


\title{Breaking the Limit of Graph Neural Networks by Improving the Assortativity of Graphs with Local Mixing Patterns}

\author{Susheel Suresh}
\affiliation{%
  \department{Department of Computer Science}
  \institution{Purdue University}
  \city{West Lafayette}
  \state{Indiana}
  \country{USA}
}
\email{suresh43@purdue.edu}

\author{Vinith Budde}
\affiliation{
    \department{Alexa AI}
    \institution{Amazon}
    \city{Seattle}
  \state{Washington}
  \country{USA}
}
\email{buddev@amazon.com}

\author{Jennifer Neville}
\affiliation{%
  \department{Department of Computer Science}
  \institution{Purdue University}
    \city{West Lafayette}
  \state{Indiana}
  \country{USA}
}
\email{neville@purdue.edu}

\author{Pan Li}
\affiliation{%
  \department{Department of Computer Science}
  \institution{Purdue University}
    \city{West Lafayette}
  \state{Indiana}
  \country{USA}
}
\email{panli@purdue.edu}

\author{Jianzhu Ma}
\affiliation{%
  \department{Department of Computer Science}
  \institution{Purdue University}
    \city{West Lafayette}
  \state{Indiana}
  \country{USA}
}
\email{ma634@purdue.edu}


\begin{abstract}
  \input{section_tex_files/abstract}

\end{abstract}

%
%

\ccsdesc[500]{Computing methodologies~Statistical relational learning}
\ccsdesc[300]{Computing methodologies~Knowledge representation and reasoning}
\ccsdesc[300]{Computing methodologies~Network science}

\keywords{graph neural networks; mixing patterns; representation learning}


\maketitle

\section{Introduction}
    \input{section_tex_files/introduction}

\section{Preliminaries}
\label{sec:prelim}
    \input{section_tex_files/preliminaries}

\section{Related Work}
\label{sec:related}
    \input{section_tex_files/related}

\section{GNNs and Local Mixing}
\label{sec:gnn_local_mixing}
    \input{section_tex_files/gnn_local_mixing}

\section{Our Framework}
\label{sec:our_framework}
    \input{section_tex_files/method}

\section{Experiments and Results}
\label{sec:exp}
    \input{section_tex_files/experiments}

    \input{section_tex_files/results}
    
    
\section{Conclusion}
    \input{section_tex_files/conclusion}

\begin{acks}
This research is supported by the National Science Foundation under contract numbers CCF-1918483, IIS-1618690, and CCF-0939370.
\end{acks}

\bibliographystyle{ACM-Reference-Format}
\bibliography{kdd}

\newpage
\appendix
\input{section_tex_files/supplemental}

\end{document}

%% file: math_commands.tex
\usepackage{amsmath,amsfonts,bm}

\def\eqref#1{equation~\ref{#1}}

\def\1{\bm{1}}

\def\va{{\bm{a}}}

\def\vh{{\bm{h}}}

\def\vm{{\bm{m}}}

\def\vp{{\bm{p}}}

\def\vx{{\bm{x}}}

\def\vz{{\bm{z}}}

\def\evy{{y}}

\def\mA{{\bm{A}}}

\def\mM{{\bm{M}}}

\def\mW{{\bm{W}}}
\def\mX{{\bm{X}}}

\DeclareMathAlphabet{\mathsfit}{\encodingdefault}{\sfdefault}{m}{sl}
\SetMathAlphabet{\mathsfit}{bold}{\encodingdefault}{\sfdefault}{bx}{n}

\def\gG{{\mathcal{G}}}

\def\emA{{A}}

\def\emM{{M}}



%% file: section_tex_files/abstract.tex
Graph neural networks (GNNs) have achieved tremendous success on multiple graph-based learning tasks by fusing network structure and node features. Modern GNN models are built upon iterative aggregation of neighbor's/proximity features by message passing. Its prediction performance has been shown to be strongly bounded by \textit{assortative mixing} in the graph, a key property wherein nodes with similar attributes mix/connect with each other. We observe that real world networks exhibit heterogeneous or diverse mixing patterns and the conventional global measurement of assortativity, such as global assortativity coefficient, may not be a representative statistic in quantifying this mixing. We adopt a generalized concept, node-level assortativity, one that is based at the node level to better represent the diverse patterns and accurately quantify the learnability of GNNs. We find that the prediction performance of a wide range of GNN models is highly correlated with the node level assortativity. To break this limit, in this work, we focus on transforming the input graph into a computation graph which contains both proximity and structural information as distinct type of edges. The resulted multi-relational graph has an enhanced level of assortativity and, more importantly, preserves rich information from the original graph. We then propose to run GNNs on this computation graph and show that adaptively choosing between structure and proximity leads to improved performance under diverse mixing. Empirically, we show the benefits of adopting our transformation framework for semi-supervised node classification task on a variety of real world graph learning benchmarks.

%% file: section_tex_files/introduction.tex
A wide variety of complex systems spanning social, chemical, technological and biological domains are modelled using graphs (networks) where nodes represent concrete or abstract entities and edges symbolize pairwise node interactions. As more data are collected, many machine learning problems on graphs emerge, such as (semi-supervised) node classification, link prediction, graph classification and graph sampling \cite{chami2020machine, ahmed2013network}, which post us new computational challenges to develop graph-based machine learning models to address these problems. In recent years, graph neural networks (GNNs) \cite{zhang2020deep_survey, hamilton2020graph} with a message passing architecture \cite{gilmer2017neural} have gained significant attention from the research community due to their promising results on various graph related ML tasks.

\begin{figure}
    \centering
    \includegraphics[width=0.45\textwidth]{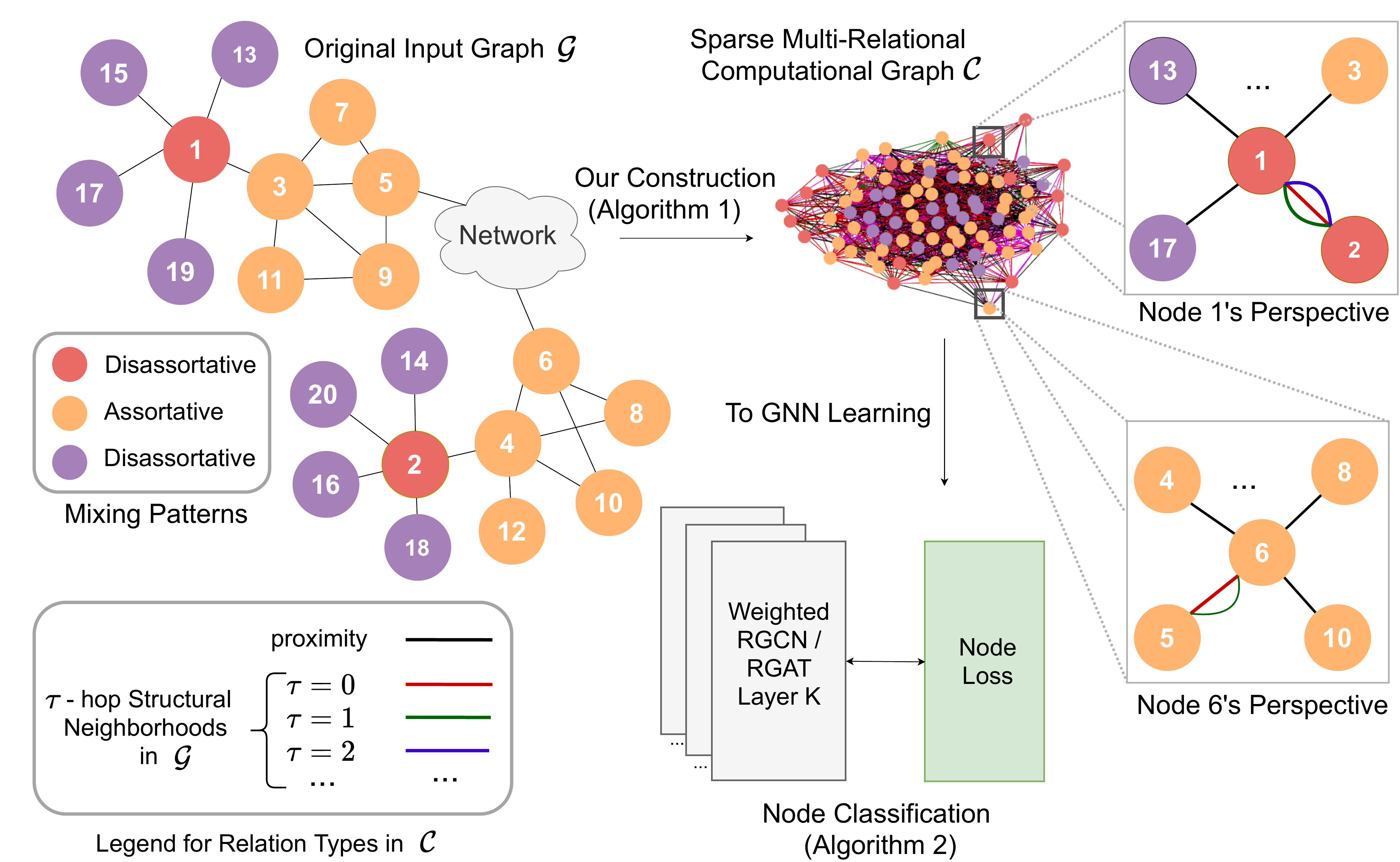}
    \caption{An input graph with diverse mixing pattern. Our pipeline uses both proximity and structural information to build a computation graph on which a GNN is run.}
    \label{fig:pipeline}
\end{figure}

In GNNs, the standard message passing works by propagating node features across edges and followed by aggregation viz. \textit{sum, mean or attention} for a number of rounds \cite{kipf2016semi, hamilton2017inductive, velivckovic2017graph}. The central idea is to utilize the neighbourhood information to construct a representation that can be beneficial for downstream learning tasks. Looking through the lens of graph signal processing, this operation of GNN could be viewed as a non-linear form of smoothing operation on the neighborhood or a low-pass graph filtering which is invariant to graph isomorphism. Clearly, one fundamental assumption made here is that similar nodes (w.r.t node attributes and labels) have a higher tendency to connect to each other compared to nodes that are far away. In other words, the philosophy followed is that proximity information from the surroundings of a node is a useful descriptor for predicting its labels. In network science, the concept \textit{assortative mixing} is defined to quantify the degree of similar node attributes/labels aggregated on local network regions \cite{mcpherson2001birds,newman2003mixing}. For instance, in social networks, people with similar habits and ideals form friendships with each other \cite{liben2007link}. In citation networks, papers from a similar area tend to cite each other. These can be recognized as \textit{assortative mixing} nodes. In the opposite, \textit{disassortative mixing} nodes can be found in technological networks where node hierarchy exists, heterosexual dating networks based on gender and ecological food webs where predator and prey tend to be dissimilar. 

In this work, we aim to study the relationships between the limits of prediction performance of GNN and different mixing patterns in a graph. Quantifying mixing patterns in graphs has conventionally been done using a global binary notion of homophily/heterophily or assortative/disassortative mixing. These global summary statistics capture the average mixing patterns in the graph as an entire entity and are meaningful only when the mixing patterns of a whole network are centered around the mean. However, most real world graphs show heterogeneous and diverse mixing patterns wherein certain parts of the graph are assortative while others disassortative \cite{peel2018multiscale,cantwell2019mixing}. For instance, Figure~\ref{fig:local_mixing_web_graph} demonstrates the distributions of assortativity on several real world graphs in which both multimodal distributions and long tail distributions are observed. Apparently, the global metrics will fail to measure this diversity on these complicated real world graphs. Recently, there has been a growing interest in designing new GNN models utilizing the nature of assortativity in graphs \cite{pei2020geom,zhu2020generalizing,liu2020non,hou2019measuring,chien2021adaptive}. However, some of the representatives are based on heuristics leveraging node attributes \cite{pei2020geom} or intermediate node representations \cite{liu2020non} to address disassortative nodes in the graph. Others simply incorporate node features from multi-hop neighbors~\cite{zhu2020generalizing} which might help improve predictions of disassortative nodes but at the same time suffer the over-smoothing problem of GNN \cite{li2018deeper, nt2019revisiting, fu2020understanding}. GPR-GNN~\cite{chien2021adaptive} may overcome the above issue using generalized graph diffusion~\cite{li2019optimizing} but loses model expressivity~\cite{chen2020graph}.

We reason that for GNNs to achieve good performance on graphs with diverse mixing, one has to provide sufficient inductive bias that lets the model adaptively choose either proximity, structural information or both for predicting node labels. This is based on the key observation that disassortative nodes (potentially far apart) may share similar structural features while assortative nodes tend to share similar features within their proximity. Consider the input graph $\mathcal{G}$ in Figure ~\ref{fig:pipeline} and two nodes colored red i.e. 1 and 2 having same labels as each other but different from the labels of their own neighbourhoods. Based on the theory of mixing patterns, these two nodes are disassortative. Even though they are far apart, their local connecting pattern is quite similar. For instance, by comparing degree sequences of nodes 1 and 2 in $\mathcal{G}$ at various neighbourhoods, we can see that at 0-hop both have similar degree of 5 and 5 respectively, their 1-hop neighbours all have the same degree of 1 or 5 and so on. This shows nodes 1 and 2 are structurally quite similar and therefore, we can make use of their structural equivalences to construct a new graph in which nodes like 1 and 2 have a connection with large weight. On the other hand, consider node like 5 and 6. They mix assortatively and their surroundings/proximity can provide enough information to infer their labels. Further, we could still benefit from the fact that nodes 5 and 6 have similar local structure. In all, our idea is to construct a transformed computation graph that encodes both structure and proximity information w.r.t each node and the GNN is run on this computation graph instead of the original graph. Note that because similar (either structurally or proximity) nodes have large weight in the computation graph it has an enhanced level of assortativity and this can boost the prediction performance of GNNs. 

To implement this idea, we first use a local measure of assortativity that can quantify diverse mixing patterns introduced in~\cite{peel2018multiscale}. This new metric, named \textit{local assortativity}, is a node-centric measure of mixing patterns that calculates assortativity within a local neighbourhood. We show that the representation capability of a wide range of GNN models is highly correlated with the level of local assortative mixing in the graph, which sets a limit to the prediction performance for GNN models based on message passing (Sec. \ref{sec:gnn_local_mixing}). To break this limit, we then develop a new algorithm which can transform the input graph into a new one with higher assortativity level and suitable for the deployment of GNN by leveraging both proximity and the local structural similarity of nodes at multiple scales. (Sec. \ref{sec:our_framework}). Figure~\ref{fig:pipeline} shows the overall framework we propose based on the idea of transforming the input graph to a new computation graph on which the GNN is run. Lastly, we conduct extensive experiments and provide analysis to show the benefits of the proposed approach (Sec. \ref{sec:exp}). Our code and an easy to use tool for evaluating GNNs w.r.t network local assortativity is provided online~\footnote{\url{https://github.com/susheels/gnns-and-local-assortativity}}.

%% file: section_tex_files/preliminaries.tex
A graph is defined as $\gG = (\mathcal{V}, \mathcal{E})$ in which $\mathcal{V}$ and $\mathcal{E}$ denote the node set and edge set, respectively. An edge going from node $u \in \mathcal{V}$ to node $v \in \mathcal{V}$ is denoted as $(u,v) \in \mathcal{E}$. The adjacency matrix $\mA \in \mathbb{R}^{|\mathcal{V}| \times |\mathcal{V}| }$ is a convenient way to represent $\gG$ where, $\mA[u,v] = 1$ if $(u,v) \in \mathcal{E}$ otherwise $0$. $\mA$ is a real valued matrix when there are \textit{weighted} edges. For \textit{multi-relational} graphs, we extend the edge notation with type as $(u, v, \tau) \in \mathcal{E}$ to denote that the edge $(u,v)$ belongs to type $\tau \in \mathcal{R}$. 
We represent the node \textit{attributes} or \textit{features} as a matrix $\mX \in \mathbb{R}^{|\mathcal{V}| \times d}$ where $d$ is the feature dimension. The feature of a particular node $u \in \mathcal{V}$ is a vector $\vx_u \in \mathbb{R}^d$. We denote the \textit{neighbourhood} around a given node $u$ which is a set of nodes exactly one hop/step away as $\mathcal{N}(u) = \{v : (u,v) \in \mathcal{E}\}$. We consider the standard semi-supervised node classification task on $\gG$, where each node $u \in \mathcal{V}$ has a class label $\evy_{u}$. The goal is to learn a function $f : \mathcal{V} \rightarrow \mathcal{Y}$ mapping the set of nodes to their class labels given some labelled nodes $\{(u_1,\evy_{u_1}), (u_2,\evy_{u_2}), \dots\}$ as training where $u_i \in \mathcal{V}$ and $\evy_{u_i} \in \mathcal{Y}$. 
\textbf{Neural Message Passing} is a framework that encompasses a range of GNN techniques inspired by the classical color refinement algorithm for graph isomorphism testing \cite{Weisfeiler1968ReductionOA,gilmer2017neural}. During this refinement process, vector messages are passed between nodes across edges and updated using neural networks repeatedly for $K$ rounds. The parameters are learned by defining a suitable loss function and followed by back propagation. A simplified and concrete working is given in the supplemental (Sec.~\ref{sec:nmp}). The global \textit{assortativity coefficient} $r_{\text{global}}$ introduced by \citet{newman2003mixing} is used to measure \textbf{Mixing in Networks} which is the tendency of nodes with similar attributes/labels to be connected to other nodes. To characterize the mixing pattern, a quantity $\emM_{gh}$ is defined to be the fraction of edges in a network that connect a node with label $g$ to one of label $h$. This helps us define a \textit{mixing matrix} $\mM$ whose elements are $\emM_{gh}$. This matrix satisfies the following sum property $\sum_g \sum_h \emM_{gh} = 1$. Global assortativity is a summary statistic for the whole network and is defined as,

\begin{equation}
    \label{eq_global_assortativity}
    r_{\text{global}} = \frac{\sum_{g} \emM_{gg} - \sum_{g} a_gb_g}{1 - \sum_{g} a_gb_g}
\end{equation}

where $a_g$ and $b_g$ represent the number of outgoing and incoming edges of all nodes of label $g$ as follows, $a_g = \sum_h \emM_{gh} $ and $b_g = \sum_h \emM_{hg}$. The quantities $a$ and $b$ can be viewed as marginals that describe the proportion of edges starting from and ending at each of the attributes. For undirected graphs where ends of edges are of same type, quantities $a_g$ and $b_g$ are equal and $\mM$ is symmetric. This allows us to write the elements of $\mM$ as,
\begin{equation}
    \label{eq_mixing_matrix}
    \emM_{gh} = \frac{1}{2m} \sum_{i:\tau_i = g} \sum_{j:\tau_j = h} \emA_{ij}
\end{equation}
where $\emA_{ij}$ is an element of adjacency matrix, $m = |\mathcal{E}|$ is the number of edges and $\tau_i$ represents the label of node $i$.

%% file: section_tex_files/related.tex
Graph Neural Networks (GNNs) have been successfully adopted for many graph related tasks \cite{duvenaud2015convolutional,allamanis2017learning, gilmer2017neural, ying2018graph, wang2019dynamic} and much effort in the community has been in understanding the nature and working of GNNs either with the lens of signal processing \cite{nt2019revisiting, fu2020understanding, min2020scattering, dong2020graph} or the combinatorial color refinement algorithm for graph isomorphism \cite{li2020distance, xu2018powerful, morris2019weisfeiler, sato2020survey}. \citet{li2018deeper} points out that GNNs essentially enforce similarity of representations between adjacent nodes akin to some sort of \textit{local smoothing}. In line with this view, \citet{nt2019revisiting} shows that GNNs behave as ``low-pass" filters filtering high frequency noise components in the convolution step. \citet{fu2020understanding} theoretically characterize the behaviour of a number of GNN models by proposing that they work by smoothing and  de-noising node features. All these results show that when node features and labels vary smoothly or in other words when there is assortative mixing, GNNs tend to work well.

Because the convolution operations are defined on neighbourhoods, the apparent local nature prohibits the use of higher-order information in the graph. To alleviate this, \citet{li2018deeper} tried to stack multiple layers of GNNs but failed due to the over-smoothing problem resulting from node representations becoming indistinguishable. This problem has also been acknowledged in \citet{klicpera2018predict}. Another line of work proposes graph attention \cite{velivckovic2017graph, hou2019measuring} computed using node features however, they are still enforcing smoothing albeit adaptively making use of relevant information from a node's surrounding. 

In light of these results, a few works propose to supplant the basic message passing framework of GNNs with extra graph information. PPNP \cite{klicpera2018predict} uses PageRank, GDC \cite{klicpera2019diffusion} utilizes graph diffusion (e.g.,heat kernels) and Geom-GCN \cite{pei2020geom} extends graph convolution with geometric aggregation derived by precomputing unsupervised node embeddings. GPR-GNN \cite{chien2021adaptive} allows different hop neighbors being associated with different signs of scalar weights to model high pass filters. Jumping Knowledge Networks~\cite{xu2018representation} leverages different neighborhood ranges for each node to enable better structure-aware representations. Non-local GNNs \cite{liu2020non} use attention to adaptively get relevant long range graph information while H$_{2}$GCN \cite{zhu2020generalizing} and MixHop \cite{abu2019mixhop} directly include information from higher order neighbourhoods within each convolution step. A comprehensive review of various graph neural networks can be found in \cite{zhang2020deep_survey,chami2020machine}.

%% file: section_tex_files/gnn_local_mixing.tex
\begin{figure}[ht]
    \centering
    \includegraphics[width=0.4\textwidth]{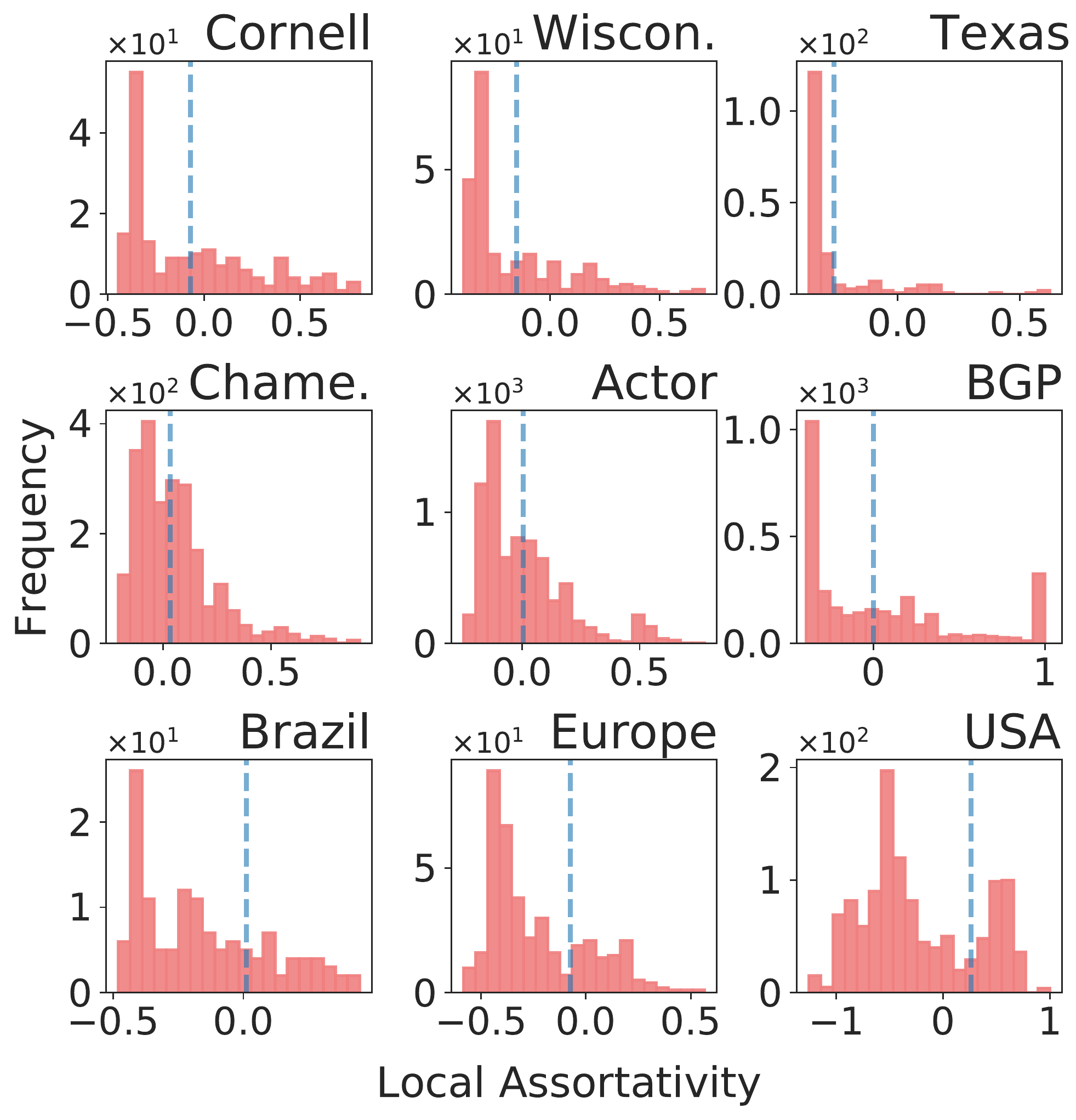}
    \caption{Observed distribution of node level assortativity in various graphs. The blue dotted line indicates global assortativity coefficient.}
    \label{fig:local_mixing_web_graph}
\end{figure}

The global assortative coefficient $r_{\text{global}}$ defined in Eq.~\ref{eq_global_assortativity} captures the average mixing pattern for the whole network but $r_{\text{global}}$ is only meaningful if all nodes have mixing concentrated around the mean. It has been studied that real world graphs exhibit high variation in mixing patterns and we are essentially interested in how GNNs perform under such a diverse mixing. For this we first utilize a node level measure of assortativity $r_{\text{local}}$ introduced by \citet{peel2018multiscale} that is calculated w.r.t a local neighbourhood. This allows us to interpolate the mixing from individual nodes to global graph level by varying the size of the local neighbourhoods. Consider a simple random walker on an undirected graph. It walks by selecting an edge at $i$ with an equal probability of $\emA_{ij}/ d_i$ where $d_i$ is the degree of node $i$. Then, the stationary probability of being at node $i$ is given by $\pi_i = d_i/2m$. This means each edge is traversed with a probability of $\pi_i \emA_{ij}/{d_i} = 1/2m$. Based on this and noting that $\emA_{ij} \in \{0,1\}$, we can rewrite Eq.~\ref{eq_mixing_matrix} as,
\begin{equation}
    \label{eq:pi_mix}
    \emM_{gh} = \sum_{i:\tau_i = g} \sum_{j:\tau_j = h} \pi_i \frac{\emA_{ij}}{d_i}
\end{equation}
Eq.~\ref{eq:pi_mix} reinterprets the mixing $\mM$ from the point of view of using random walk to visit the entire graph and thus reveals that the global assortativity counts all edges in the graph equally. For the local measure of assortativity the edges are weighted according to how local they are to a node of interest $l$ by replacing the stationary distribution $\pi_i$ with an alternative distribution over the nodes $w(i;l)$. Then Eq. \ref{eq:pi_mix} becomes,
\begin{equation}
    \label{eq:mixing_with_dist}
    \emM_{gh}(l) = \sum_{i:\tau_i = g} \sum_{j:\tau_j = h} w(i;l) \frac{\emA_{ij}}{d_i}
\end{equation}
The personalized PageRank vector is utilized as a proxy for $w(i;l)$. Concretely, it is simple random walk with restarts i.e. during a simple random walk, the walker can return to the initial node of interest $l$ with a probability of $(1-\alpha)$. Varying $\alpha$ allows us to interpolate from the trivial local neighbourhood (when $\alpha = 0$ i.e. walker never leaves the node) to the global assortativity (when $\alpha$ = 1 i.e. no restarts). Finally, the local assortativity metric for a node $l$ parameterized by $\alpha$ is,
\begin{equation}
    \label{eq_local_assortativity}
    r_{\alpha}(l) = \frac{\sum_{g} \emM_{gg}(l) - \sum_{g} a_g^2}{1 - \sum_{g} a_g^2}
\end{equation}
Note that we can recover the global assortativity metric from Eq.~\ref{eq_local_assortativity}, $r_{1}({l}) = r_{\text{global}}$ because when ($\alpha = 1$)  no restarts happen, $w_{\alpha}(i;l)$ falls back to $\pi_{i}$. When calculating the $r_{\alpha}(l)$ in practice, instead of choosing $\alpha$ heuristically, inspired by TotalRank \citep{boldi2005totalrank}, the  PageRank vector is averaged over the entire range of $\alpha \in [0,1]$ as,
\begin{equation}
    w_{\text{tt}}(i;l) = \int_0^1 w_{\alpha} (i;l) d\alpha
\end{equation}
With $w_{tt} (i;l)$ in place of $w(i;l)$ in Eq.~\ref{eq:mixing_with_dist} and finally using the resultant mixing matrix in Eq.~\ref{eq_local_assortativity} gives us our local assortativity metric $r_{\text{local}}(l)$.

In Figure~\ref{fig:local_mixing_web_graph}, we examine various networks from different domains for existence of diverse mixing patterns using $r_{\text{local}}$. The details of these networks are given in the supplemental (Sec.~\ref{sec:dataset}). In almost all of them we witness skewed and multimodal distributions. It is interesting to note that the global assortativity coefficient $r_{global} \approx 0$ (Eq.~\ref{eq_global_assortativity}) while nodes exhibit diverse mixing over the full spectrum of $r_{\text{local}}$. This observation highlights that $r_{global}$ is not necessarily reliable as it doesn't give a complete picture.

\begin{figure}
    \centering
    \includegraphics[width=0.47\textwidth]{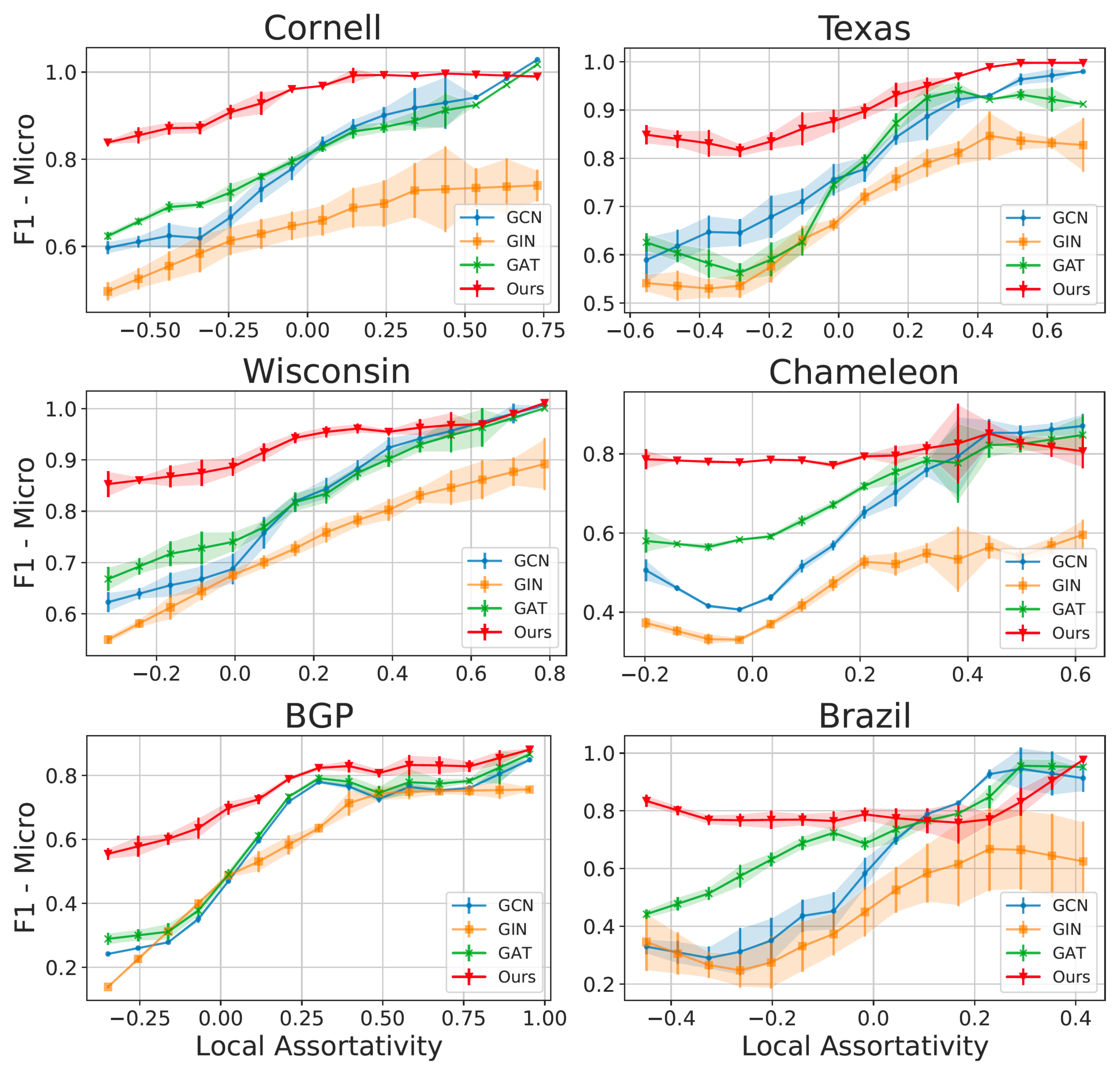}
    \caption{GNNs vs Local Assortativity on various graphs. Node classification is performed on various datasets (See Table~\ref{tab:dataset_statistics} for details). $Y$-axis shows mean F1-Micro with standard deviation over 10 runs.}
    \label{fig:gnn_vs_local}
\end{figure}

\subsection{Analysis of GNNs Under Local Mixing}
Given the observation in Figure~\ref{fig:local_mixing_web_graph}, we are interested in studying the behaviour of various leading GNN models such as GCN \cite{kipf2016semi}, GIN \cite{xu2018powerful} and GAT \cite{velivckovic2017graph}, when applied to graphs with diverse mixing patterns. The task of semi-supervised node classification is used as a proxy to understand the power of modelling graph data w.r.t different levels of local assortativity embedded in the graph. As shown in Figure~\ref{fig:gnn_vs_local}, the performance of GNN models are highly correlated with the node-level local assortativity $r_{\text{local}}$ within the same graph they are deployed. Across all the tested real world graphs, another clear pattern is that most of the popular GNN methods perform poorly for disassortative nodes $l$ with $r_{\text{local}} (l) < 0$. The reason is that the features of disassortative nodes are vastly different from their neighbourhoods' and, GNN methods simply cannot create useful node representations based on the information provided by their neighbours through conventional node smoothing operation (alluded in theoretical works \cite{nt2019revisiting, fu2020understanding}). We further characterize our reasoning below with the help of two definitions. 

\theoremstyle{definition}
\begin{definition}[Neighbourhood Label Smoothness]
\label{def:nls}
A node $u \in \mathcal{V}$ with class label $\evy_u \in \mathcal{Y}$ has label smoothness parameter defined on the neighbourhood $\mathcal{N}(u)$ as, $\epsilon_u = \frac{1}{|\mathcal{N}(u)|}\sum_{i \in \mathcal{N}(u)} P(\evy_{i} = \evy_{u}| \evy_u)$
\end{definition}

\theoremstyle{definition}
\begin{definition}[Neighbourhood Feature Smoothness]
\label{def:nfs}
A node $u \in \mathcal{V}$ with feature vector $\vx_u \in \mathbb{R}^d$ has a smoothness parameter defined on the neighbourhood $\mathcal{N}(u)$ as, $\lambda_u = || \vx_u - \frac{1}{|\mathcal{N}(u)|}\sum_{i \in \mathcal{N}(u)} \vx_i ||^2$
\end{definition}

Labels of disassortative nodes and their neighbours have high probability of being different which means their labels tend to be not smooth in that neighbourhood, which is quite clear from Def.~\ref{def:nls} (their $\epsilon_u$'s tends to be low). Then, it is also likely that features of such nodes are not smooth either (high $\lambda_u$). This is based on a reasonable assumption that features $\vx_u$ and class labels $\evy_u$ are correlated. However, GNNs try to smooth the node features in the latent space which in turn smooth out the class label predictions. Thus, performing poorly in a disassortative regime. On the flip side for neighbourhoods mixing assortatively, owing to smooth labels all across, $\epsilon_u$ will be large, which also means high feature smoothness (low $\lambda_u$), a regime that is very beneficial to GNNs. A key take away is that $\epsilon_u \propto 1/\lambda_u$ and this characterization explains the observations we witness in Figure~\ref{fig:gnn_vs_local}.

%% file: section_tex_files/method.tex
In Section~\ref{sec:gnn_local_mixing} we showed that real world graphs contain diverse mixing patterns and experimental results demonstrate that current GNN methods based on message passing are unable to provide good representations for nodes that show disassortative mixing. To break the limit introduced by the original graph structure, we design a graph transformation algorithm which can generate a new graph with higher assortativity. The central idea is to leverage the structural and proximity information in the original graph. Ideally, the resulted new graph should have the following properties: (1) Constructed from the original input graph using the same set of nodes without the class label information. (2) Encode the structural equivalences in the original input graph in a model free manner. (3) Encode proximity information as seen in the original graph.

These properties requires the construction of a new graph to be solely based on structural and proximity regularities of the original graph. Traditionally, GNN methods define convolutions on the original input graph and various design choices are made. Different from them, our framework aims to apply such GNN models on a transformed graph which we dub as the ``computation graph" for the same machine learning task. Note that our framework consists of 2 stages. Stage (1) transforms the original input graph to a computation graph and in Stage (2) GNN methods are applied on the computation graph for learning. In the next section, we introduce one approach for graph transformation and later define message passing on the transformed graph.

\subsection{Computation Graph }
One obvious choice to encode structural equivalences between nodes is to compare ordered degree sequences at various hierarchies \cite{ribeiro2017struc2vec}. The rationale is that any two nodes with same degree are structurally similar, and if their one hop neighbours also have same degree, then they are even more structurally similar and so on. Therefore, a key observation is that the structural similarity between two nodes monotonically increases when their degree sequences get progressively similar. More formally, let $\mathcal{N}_\tau(g)$ denote the set of neighbouring nodes at exactly $\tau$ hops away from node $g$ in graph $\mathcal{G}$. Let $s(V)$ represent a non-increasing (ordered) sequence of degrees of a set $V \subset \mathcal{V}$ of nodes. The goal is to compare ordered degree sequences at various neighbourhoods for every pair of nodes $(g, h)$ in $\mathcal{G}$. The notion of \textit{structural distance} \cite{ribeiro2017struc2vec} is recursively defined as follows,
\begin{equation}
    \label{eq:sd}
    \begin{aligned}
    f_{\tau}(g,h) = f_{\tau-1}(g,h) + \mathcal{D} (s (\mathcal{N}_\tau(g)), s (\mathcal{N}_\tau(h))) 
  \end{aligned}
\end{equation}
where $\mathcal{D}(S_1, S_2) \geq 0$ measures the distance between ordered degree sequences $S_1$ and $S_2$ and $f_{-1}(\cdot) = 0$. $f_\tau(g,h)$ is defined for $\tau \geq 0$ and $|\mathcal{N}_\tau(g)|, |\mathcal{N}_\tau(h)| > 0$ i.e. only when neighbourhoods at $\tau$ exist. The cost function $\mathcal{D}(\cdot, \cdot)$ should ideally give small values for similar ordered degree sequences while provide large values for vastly different ones. Following \cite{ribeiro2017struc2vec}, we make use of Fast Dynamic Time Warping (DTW) \cite{salvador2007fastdtw} that is best suited for loosely comparing sequences of different sizes. The recursive definition of structural distance in Eq.~\ref{eq:sd}, makes sure that $f_\tau(\cdot, \cdot)$ can only increase as we successively progress through $\tau$. So, for nodes $g$ and $h$, that are structurally similar, structural distance is low. 

\makeatletter 
\g@addto@macro{\@algocf@init}{\SetKwInOut{paramH}{Hyper-Params.}} 
\makeatother
\begin{algorithm}
\caption{Construction of computation graph}
\label{algo:construction_algorithm}
    \KwIn{Original Input Graph $\mathcal{G} = (\mathcal{V}, \mathcal{E})$; 
    }
    \paramH{No. of structural relations $T < dia(\mathcal{G})$}
    \KwOut{Computation graph $\mathcal{C}$}
     \Begin{
        
        $\mathcal{E}^\prime \gets \emptyset $;$\; n \gets |\mathcal{V}|$\;
        $ w_{\tau} \in \mathbb{R}^{n \times n} \gets 0, \forall \tau \in \{0,1 \cdots T\}$;$\; w_{p} \in \mathbb{R}^{n \times n} \gets 0$\;
    
        $f_{-1} \gets 0$\;

        \For{$(g,h) \in \mathcal{V}^2$}{
             \For{$\tau \in \{0,1 \cdots T\}$}{
                
                $f_{\tau}(g,h) \gets f_{\tau-1}(g,h) + \mathcal{D} (s (\mathcal{N}_\tau(g)), s (\mathcal{N}_\tau(h)))$\;
                
                
                $w_\tau(g,h) \gets e^{-f_\tau(g,h)}$ 
                
                
                
                $\mathcal{E}^\prime \gets \mathcal{E}^\prime \cup (g,h,\tau)$\;

            }
        }
        \For{$(g,h) \in \mathcal{E}$}{
            $\mathcal{E}^\prime \gets \mathcal{E}^\prime \cup (g,h,p)$; $\;w_{p} (g,h) \gets 1$
            
        }
        $\mathcal{R} \gets \{0,1 \cdots T\} \cup p$\;
        \Return{$\mathcal{C} = \big(\mathcal{V}, \mathcal{E}^\prime, \mathcal{R}, \{w_{\tau}, \forall \tau \in \mathcal{R}\}\big)$}
    }

\end{algorithm}

\subsubsection{Graph construction} 
We construct a weighted multi-relational computation graph $\mathcal{C}$ which encodes the structural distances at various hierarchies between pairs of nodes and proximity information available in the original input graph $\mathcal{G}$. Let $T$ denote a number less the diameter of the graph $\mathcal{G}$ and $n = |\mathcal{V}|$ the number of nodes. To construct the new graph $\mathcal{C}$, we first add the original node set $\mathcal{V}$ and for each pair of nodes $(g, h)$ we create $T~+~1$ different types of edges where each type corresponds to the $\tau$-hop neighbourhoods on which we calculated structural distance in $\mathcal{G}$. To encode structural equivalence between nodes we define edge weights $w_\tau(g,h)$ between $g$ and $h$ with structural relation type $\tau$ to vary inversely with structural distance $f_\tau(g,h)$ as follows:
\begin{equation}
\label{eq:weight}
    w_\tau(g,h) = e^{-f_\tau(g,h)}, \;\;\; \tau = 0, 1, \dots T
\end{equation}
The edge weight for type $\tau$ between $g$ and $h$ is large when their $\tau$-hop neighbors have similar network structure properties (low structural distance). 

For proximity information, we simply use the original edges of $\mathcal{G}$ and add it to $\mathcal{C}$ with weight one.
In total, this construction creates $\mathcal{C}$ which has $\mathcal{V}$ nodes and at most $\mathcal{E} + (T + 1) \genfrac(){0pt}{2}{n}{2}$ edges. The algorithm for the construction of this new graph is detailed in Algorithm~\ref{algo:construction_algorithm}. Naively using the above graph construction results in a large number of edges being introduced. For practicality, pairwise structural similarity calculations are restricted to $O(\log n)$ for each node at each $\tau$-hop neighbourhood. We thus have $O(n \log n)$ edges for each $\tau$ instead of $O(n^2)$. For completeness we provide the efficient practical implementation in the supplement (Sec. \ref{algo:construction_algorithm_practical}). 

In this work, we provide one specific implementation of our general idea of using both structure and proximity information from the original graph into the computation graph. Other structural techniques viz. RolX \cite{henderson2012rolx}, GraphWave~\cite{donnat2018learning} and generalized proximity inspired methods viz. Graph Diffusion~\cite{klicpera2019diffusion}, PageRank~\cite{klicpera2018predict} can also be adopted. 

\subsection{Message Passing on the Multi-relational Computation Graph} 
The constructed multi-relational graph $\mathcal{C}$ and the original graph $\mathcal{G}$ share the same node set $\mathcal{V}$ and can be defined as $(\mathcal{V}, \mathcal{E}^\prime, \mathcal{R})$. Each edge in $\mathcal{C}$ going from node $u$ to $v$ of type $\tau$ is represented as a triplet $(u,v,\tau) \in \mathcal{E}^\prime$. Recall that there are $T+1$ structural type edges and one proximity type edge. Thus, the total number of relations $|\mathcal{R}| = T + 1 + 1$. We now define the message passing procedure on $\mathcal{C}$ following the standard GNN formulation (Sec.~\ref{sec:nmp} ).

To account for the different relation types in $\mathcal{C}$, following \cite{schlichtkrull2018modeling}, we introduce a relation specific transformation matrix $\mW_{\tau}$ for each type $\tau \in \mathcal{R}$ of edge and specify the AGGREGATE function as follows,
\begin{equation}
\label{eq:aggregate}
    \vm_{u} = \sum_{\tau \in \mathcal{R}} \sum_{v \in \mathcal{N}_{1}^{\tau}(u)} \mW_{\tau} \; \vh_{v} \; w_{\tau}(u,v) \; \alpha_{\tau}(u,v)
\end{equation}
where $w_{\tau}(u,v)$ is the $\tau$ type specific edge weight between nodes $u$ and $v$ defined in Algo.~\ref{algo:construction_algorithm}. In line with our motivation of letting the model adaptively choose between structural and proximity information, we make use of an attention mechanism \cite{velivckovic2017graph} defined using attention coefficients $e_{\tau}(u,v)$ that indicates the importance of node $v$'s feature to node $u$ w.r.t a particular relation type $\tau$.
\begin{equation}
    e^{\tau}_{u,v} = a_{\tau}(\mW_{\tau} \vh_{u}, \mW_{\tau} \vh_{v} ) = \va^{T}[\mW_{\tau} \vh_{u} || \mW_{\tau} \vh_{v} ]
\end{equation}
where $ a : \mathbb{R}^d \times \mathbb{R}^d \rightarrow \mathbb{R}$ is a shared attention mechanism parameterized by a learnable attention weight vector $\va \in \mathbb{R}^{2d}$, $\cdot^{T}$ is transpose operation and $||$ is concatenation operation. We inject the available computation graph information as follows,
\begin{equation}
\label{eq:atten}
    \alpha_{\tau} (u, v) = \frac{\text{exp}(\text{LeakyReLU}(e^{\tau}_{u,v}) )}
    {\sum\limits_{x \in \mathcal{N}_{1}^{\tau}(u)} \text{exp}(\text{LeakyReLU}(e^{\tau}_{u,x}))}
\end{equation}
Finally, The UPDATE function is defined as, 
\begin{equation}
    \vh_u^{\prime} = \sigma \Big(\mW_{\text{self}} \; \vh_u + \; \mW_{\text{neig}} \; \vm_{u}    \Big)
\end{equation}
We name our general model as WRGNN (weighted relational GNN) and from the above definitions, we select two model variants viz. WRGAT and WRGCN for experimental analysis which specifies if attention mechanism is used or not respectively. Algorithm~\ref{algo:our_algorithm} provides a procedure for applying such a $K$ layer WRGNN on the computation graph for semi-supervised node classification.

%% file: section_tex_files/experiments.tex
In this section, we evaluate the performance of our framework against other methods under semi-supervised node classification setting. Note that our framework is quite flexible so any GNN model based on message passing could be adopted on our computation graph. To evaluate the performance of our method, we use real world graphs from different domains viz. Hyperlinked Web Pages~\cite{pei2020geom}, Citation Networks~\cite{namata2012query}, Air Traffic Networks~\cite{ribeiro2017struc2vec} and Internet's Inter-Domain Routing Network~\cite{luckie2013relationships, hou2019measuring}. These graphs are know to exhibit diverse mixing as we showed in Sec.~\ref{sec:gnn_local_mixing} and thus provides us with the means to assess GNN based methods. Statistics and brief descriptions are provided in Table~\ref{tab:dataset_statistics} and Sec.~\ref{sec:dataset}.

\subsection{Baseline Methods and Experiment Setup}
\label{sec:exp_setup}
We primarily consider methods that utilize GNN models which adopted messaging passing operation as their main backbones. GCN \cite{kipf2016semi} and GraphSage \cite{hamilton2017inductive} are methods where convolutions are strictly based on first order neighbour aggregation scheme for each layer. GCN-Cheby \cite{defferrard2016convolutional} generalizes convolutions with the help of k-hop localized spectral filters. GAT \cite{velivckovic2017graph} adaptively aggregates immediate neighbour information using attention coefficients which are also derived from node features. MixHop \cite{abu2019mixhop} and H$_2$GCN \cite{zhu2020generalizing} generalize the node aggregation beyond the first order neighbourhoods and dynamically considers node features $k$-hops away. It is important to note that all baseline methods operate on the original graph and thus have access to only proximity information albeit in different forms.

We perform semi-supervised node classification and use the classification accuracy and F1-Micro scores as performance metrics to evaluate different approaches. The training/validation/testing data splits for all the methods to be compared is shown in Table~\ref{tab:dataset_statistics}. For Hyperlinked Web Page and Citation Networks, we report the performance of mean $\pm$ std. dev. on 10 random splits provided by Pei et al. \cite{pei2020geom} which is available on their GitHub~\footnote{https://github.com/graphdml-uiuc-jlu/geom-gcn/tree/master/splits}. 
Reported values for Air Traffic Networks and BGP Networks are based on 20 and 10 random splits respectively. All the implemented methods including our own are all trained until the loss function converges and the final models are selected based on the prediction performance on the validation sets. The sensitivity analysis and hyper-parameter search is performed on the validation set and more details are provided in the supplement (Sec.~\ref{hyper_setting}).


%% file: section_tex_files/results.tex
\begin{figure}[htb]
    \centering
    \includegraphics[width=0.4\textwidth]{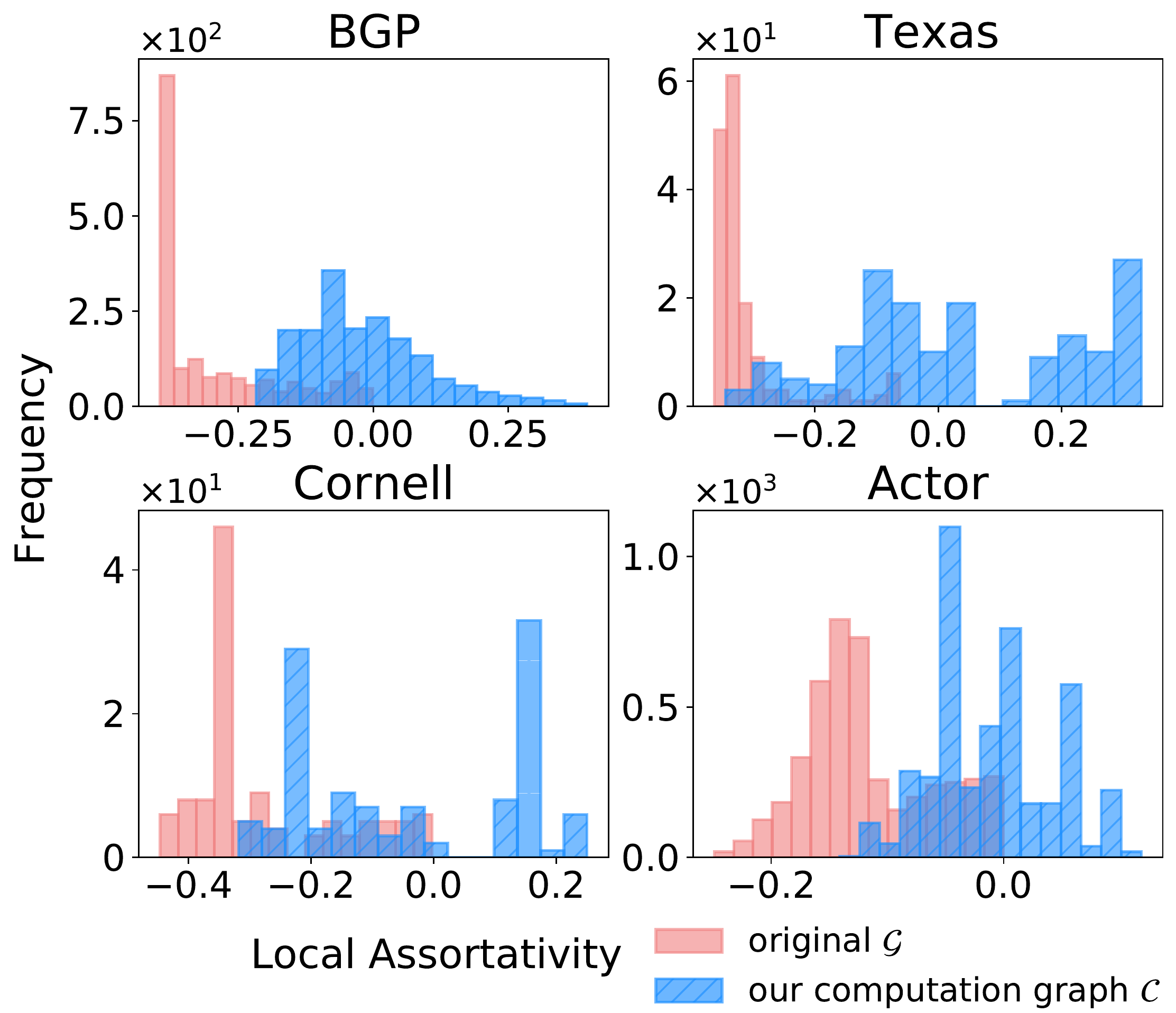}
    \caption{Distributions of node level assortativity for the original graph and the newly constructed computation graph on different datasets.}
    \label{fig:distribution_shift}
\end{figure}

\subsection{Local Assortativity Distribution Shift}
We first perform a quick study to confirm our graph transformation algorithm indeed enhances the level of local assortativity in comparison to the original input graph. To achieve this, we focus on all the disassortative nodes ($r_{\text{local}} < 0$) from the input graph $\mathcal{G}$ and track how they mix in the transformed computation graph $\mathcal{C}$. From Figure~\ref{fig:distribution_shift}, a clear distribution shift of local assortativity could be observed. That is, the previously disassortative nodes in $G$, are more assortative in the new computation graph after transformation. This empirically verifies the claim we raised earlier about similar structural regularities between a pair of nodes being captured in the computation graph as a result of our transformation (hence the increased assortativity). In addition, we have a reason to believe that performance improvement of our relational GNN model variants WRGCN and WRGAT are rooted in the increase of local assortativity for disassortative nodes in the original graph. This is clearly witnessed in Figure~\ref{fig:gnn_vs_local} (shown in red).


\subsection{Performance of Node Classification}

\input{tables/main_results}

\input{tables/airports_results}

\noindent \textbf{Hyperlinked Web Page and Citation Networks.} Table~\ref{tab:main_results} shows the evaluation of our framework against other GNN based baselines on the node classification task for these datasets. Mean test accuracy and standard deviation numbers are reported for each method. Values for H$_2$GCN \cite{zhu2020generalizing}and Geom-GCN \cite{pei2020geom} are taken from the respective papers. We use the same train/val/test splits as \cite{zhu2020generalizing,pei2020geom} for comparability. We find that our framework consistently performs well for the Hyperlinked Web Page Networks owing to the rich structural regularities that our computation graph captures. On Citation Networks which predominantly have assortative mixing, our framework gives comparable performance to baselines. H$_2$GCN and MixHop utilize higher order neighbourhoods in each convolution which does help over standard GNN methods, but can also be a lot harder to train and they also suffer from oversmoothing problem. Our framework makes it possible to define computation graphs that can directly tap into the structural regularities thereby making effective use of graph information, while the message passing defined on such a computation graph takes care of adapting to node features.
\begin{figure}
    \centering
    \includegraphics[width=0.45\textwidth]{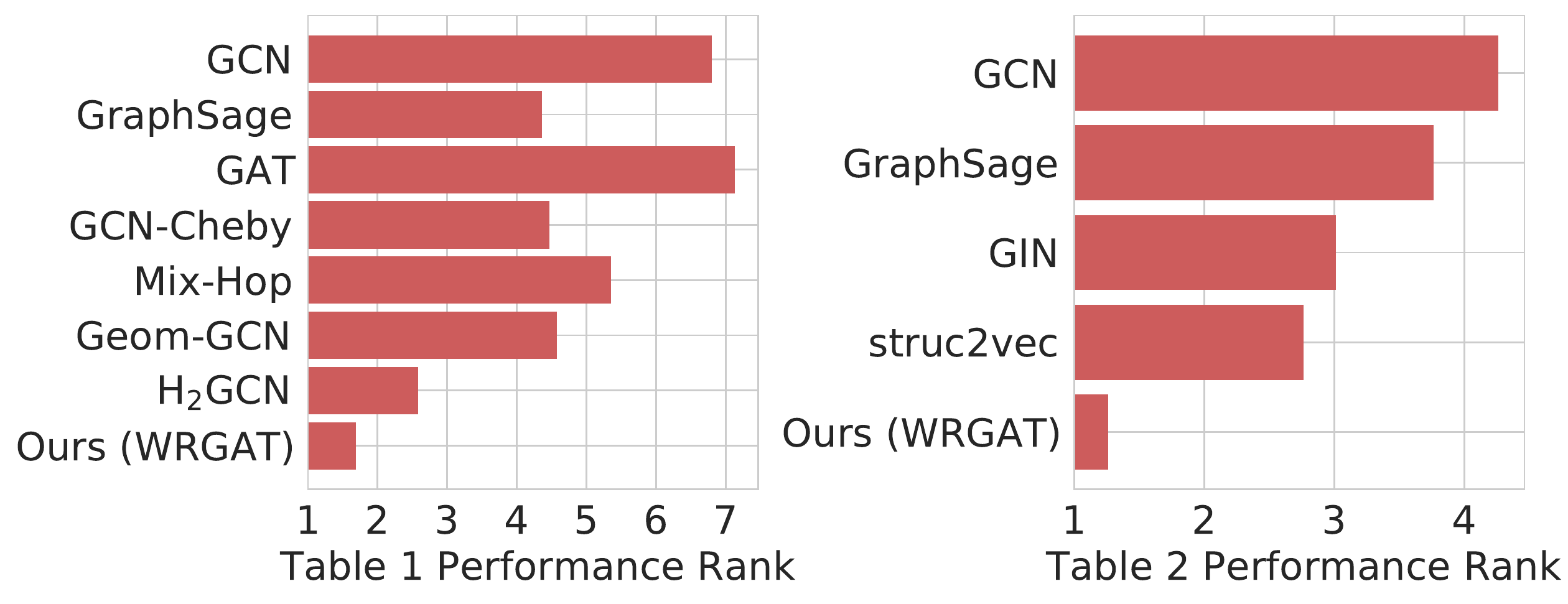}
    \caption{Overall perf. rank of various methods for datasets in Table \ref{tab:main_results} \& \ref{tab:airports_bgp_results}. Lower rank signifies better performance}
    \label{fig:perf_rank}
\end{figure}
Overall performance rank of various methods in Table~\ref{tab:main_results} on both disassortative and assortative datasets is shown in Figure~\ref{fig:perf_rank}. It is clear that our model variant WRGAT run on the computation graph achieves the lowest rank overall hence supporting our claim of achieving superior performance in both disassortative and assortative regime.

\noindent \textbf{Air Traffic and BGP Networks.} Table~\ref{tab:airports_bgp_results} gives comparisons of our framework against other GNN methods for node classification on multiple Air Traffic Networks (ATNs) and the Internet Domain Network (BGP). It is clear from the table that, our framework achieves strong performance compared to other baselines. ATNs don't have node attributes and baseline GNN methods perform poorly while our framework utilizing structure is better suited for the task. Strong performance is seen because our construction of the computation graph explicitly looks at different neighbourhoods that is very beneficial in air traffic networks. Major hubs connect to local airports (disassortative) while they also mix with other hubs (assortative). Structure alone is capable of capturing this diversity and our computation graph takes a step in that direction. The BGP network also exhibits diverse mixing and we believe that the strong performance is due to the adaptive selection of both proximity and structural information. 

\subsection{Sensitivity Analysis}
\begin{figure}
    \centering
    \includegraphics[width=0.4\textwidth]{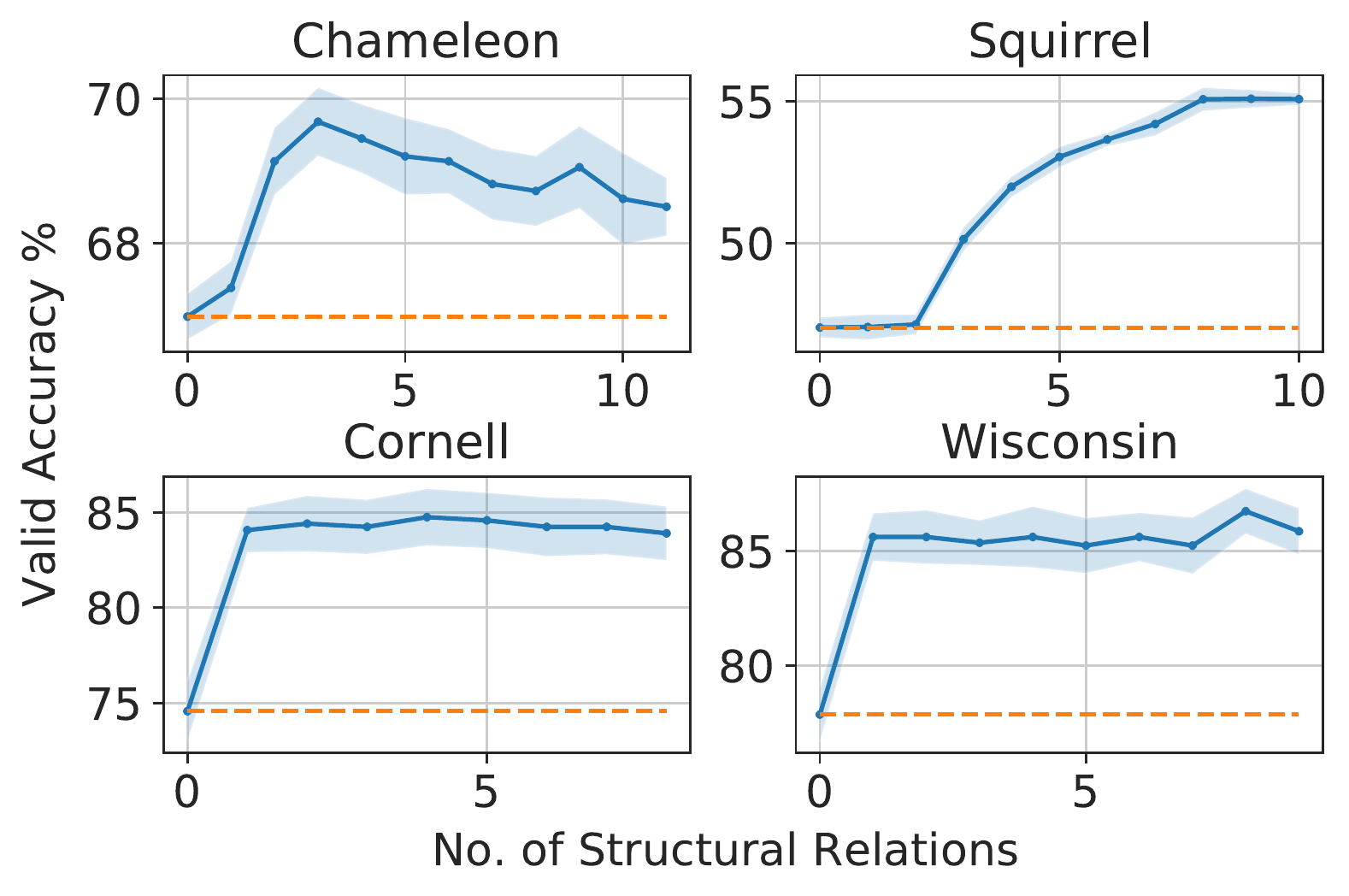}
    \caption{Sensitivity w.r.t structure relations. Baseline is using only proximity information.}
    \label{fig:sensitivity_structure_relations_disassortative}
\end{figure}

\begin{figure}
    \centering
    \includegraphics[width=0.45\textwidth]{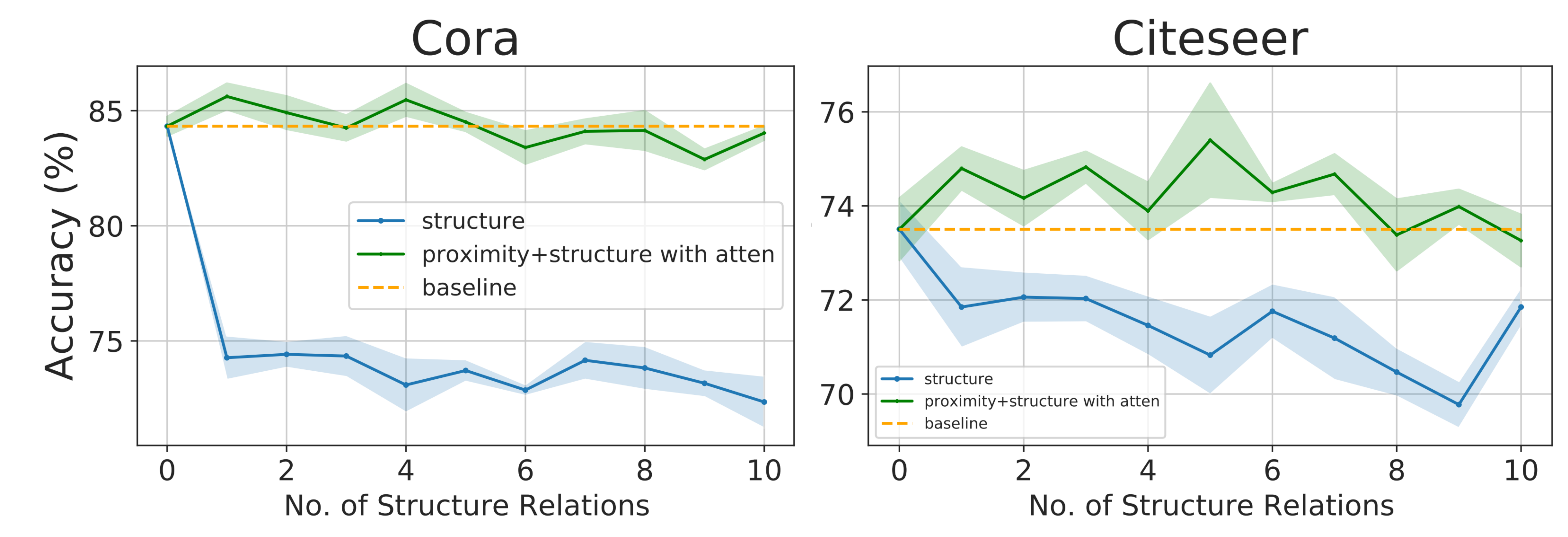}
    \caption{Sensitivity w.r.t structure relations. Baseline is using only proximity information.}
    \label{fig:sensotivity_structure_relations_cora_citeseer}
\end{figure}
We provide sensitivity analysis based on validation split w.r.t the number of structural relations used in our computation graph, which is our main hyper-parameter. Figure~\ref{fig:sensitivity_structure_relations_disassortative} shows the analysis for disassortative networks and is quite clear that adding structural information at various scales significantly improves validation performance when compared to baseline of just using proximity information. Figure~\ref{fig:sensotivity_structure_relations_cora_citeseer} shows an interesting picture for highly assortative networks. Here, the take away is that structure alone is not useful (blue line), while an adaptive structure+proximity with attention model i.e. using WRGAT (green line) is able to give better consistent performance over proximity only baseline.

\subsection{Ablation Analysis}

\begin{figure}
    \centering
    \includegraphics[width=0.5\textwidth]{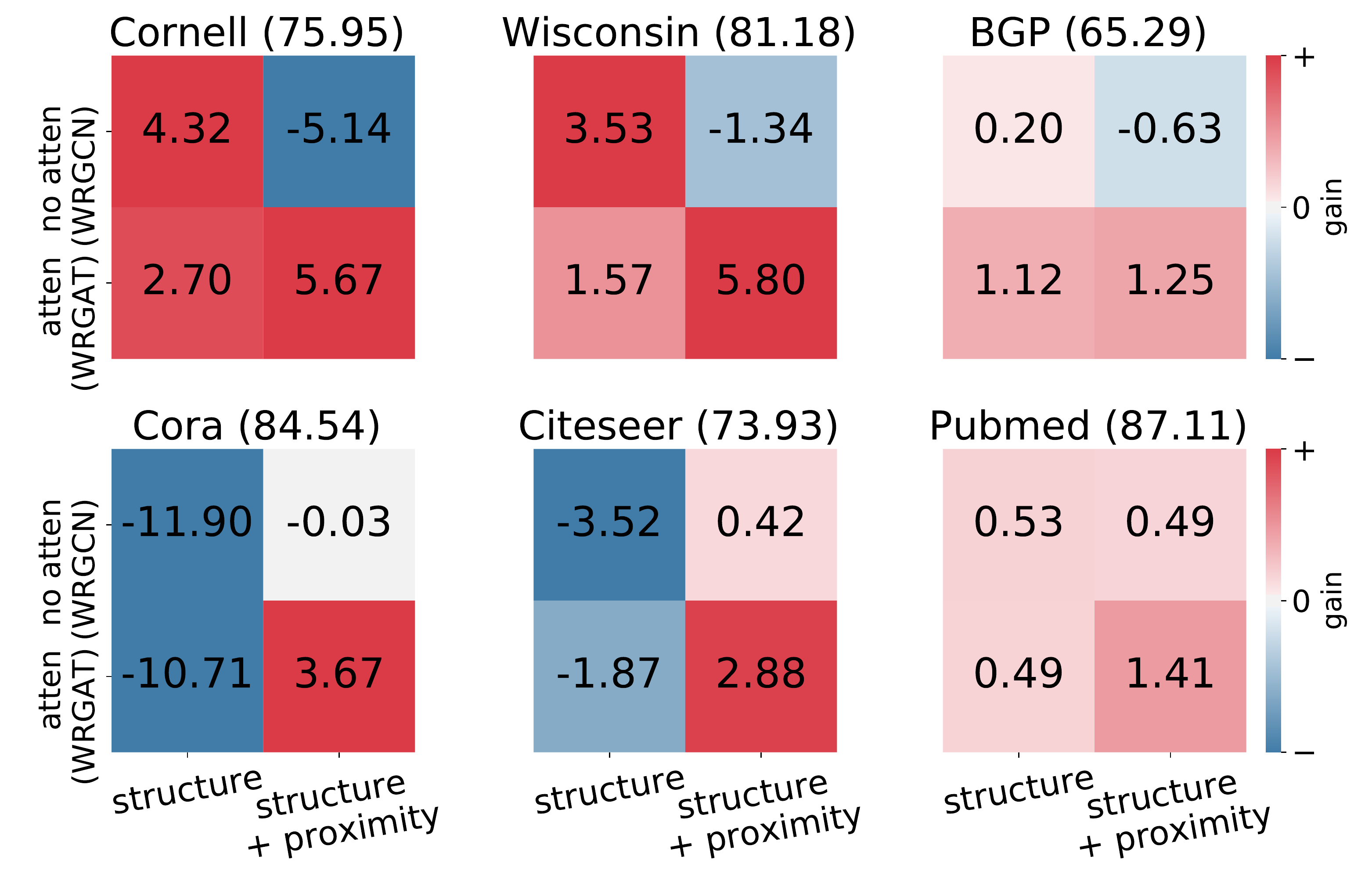}
    \caption{Ablation w.r.t different model and graph information choices. Baseline (proximity only) is shown in parenthesis. Numbers indicate gain over baseline.}
    \label{fig:ablation_analysis}
\end{figure}

\begin{figure}
    \centering
    \includegraphics[width=0.45\textwidth]{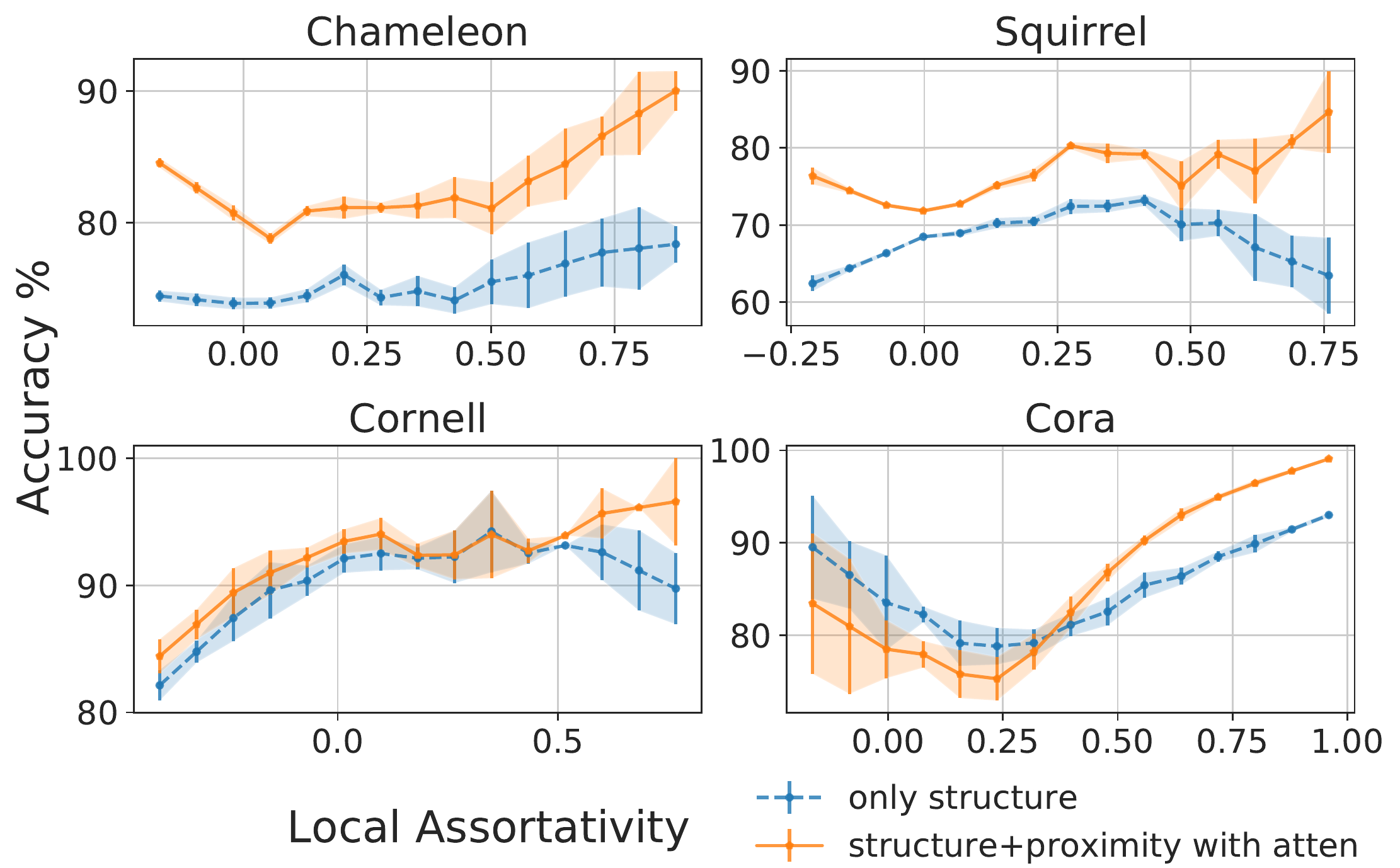}
    \caption{Ablation w.r.t graph info and local assortativity}
    \label{fig:ablation_analysis_local_assortativity}
\end{figure}
Figure~\ref{fig:ablation_analysis} provides ablation analysis for a number of networks and shows test performance gains over proximity only baseline for our model variants (WRGCN and WRGAT) in y-axis and against the use of either proximity, structure or both kinds of information in x-axis. For the top row consisting of predominantly disassortative networks, the take away is that structure only information is capable of providing better performance over proximity only information, but when both kinds of information is available, the attention model is best suited. For the bottom row consisting of highly assortative networks, clearly structure only information hurts performance compared to baseline. We reason that structure becomes less important for graphs with high assortativity however, when both kinds of information is available and attention is used we witness the best gains. These results further supports our model choices.

The previous ablation study shows test performance gains for the whole dataset. We now study how these different model configurations behave w.r.t node level local assortativity we introduced earlier. With Figure~\ref{fig:ablation_analysis_local_assortativity} we are able to study their behaviour in finer detail due to the notion of local assortativity. It supports our hypothesis that in the high local assortativity regime, proximity information dominates in discriminatory power and structure only information leads to worsened performance as it becomes irrelevant. However when using both kinds with an adaptive mechanism we can see increased performance over the full spectrum.

%% file: tables/main_results.tex
\begin{table*}[htb]
\centering
\renewcommand{\arraystretch}{0.95} 

\caption{Semi-supervised node classification showing mean test accuracy $\pm$ std. over 10 runs. Club Suit [$\clubsuit$] denotes result obtained from the best model variant of respective papers.}
\label{tab:main_results}
\begin{tabular}{@{}lllllll|llll@{}}
\toprule
            & Chameleon & Squirrel & Actor & Cornell & Texas & Wisconsin & Cora & Citeseer & Pubmed \\ \cmidrule(l){2-10} 

GCN         &  59.82 $\pm$ 2.58          &  36.89$\pm$1.34        &  30.26$\pm$0.79     & 57.03$\pm$4.67        & 59.46$\pm$5.25      & 59.80$\pm$6.99          &   87.28$\pm$1.26   & 76.68$\pm$1.64   & 87.38$\pm$0.66       \\

GraphSage   &  58.73$\pm$1.68         &  41.61$\pm$0.74         &   34.23$\pm$0.99    & 75.95$\pm$5.01        & 82.43$\pm$6.14      &  81.18$\pm$5.56   &  86.90$\pm$1.04    &  76.04$\pm$1.30        & 88.45$\pm$0.50       \\

GAT         &  54.69$\pm$1.95         &  30.62$\pm$2.11        &  26.28$\pm$1.73     &     58.92$\pm$3.32    &  58.38$\pm$4.45     &     55.29$\pm$8.71      &    86.37$\pm$1.69  &    75.46$\pm$1.72      &    87.62$\pm$0.42    \\

GCN-Cheby   &   55.24$\pm$2.76        & 43.86$\pm$1.64         &    34.11$\pm$1.09   & 74.32$\pm$7.46        &  77.30$\pm$4.07     &    79.41$\pm$4.46       & 86.86$\pm$0.96     &     76.25$\pm$1.76     &    88.08$\pm$0.52    \\

MixHop      & 60.50$\pm$2.53          & 43.80$\pm$1.48         &    32.22$\pm$2.34   & 73.51$\pm$6.34        & 77.84$\pm$7.73      &    75.88$\pm$4.90       &  83.10$\pm$2.03    & 70.75$\pm$2.95         &    80.75$\pm$2.29    \\

Geom-GCN $\clubsuit$  &  60.90         & 38.14         & 31.63      & 60.81        & 67.57      &     64.12      &    85.27  &    77.99      &    90.05    \\

H$_2$GCN $\clubsuit$    &    59.39$\pm$1.98       & 37.90$\pm$2.02         & 35.86$\pm$1.03      &  82.16$\pm$4.80       &  84.86$\pm$6.77     & 86.67$\pm$4.69          &  87.67$\pm$1.42    &  76.72$\pm$1.50        & 88.50$\pm$0.64       \\ \midrule

Ours (WRGAT) &  65.24$\pm$0.87         &   48.85$\pm$0.78       &  36.53$\pm$0.77     &    81.62$\pm$3.90     &   83.62$\pm$5.50    &   86.98$\pm$3.78        &  88.20$\pm$2.26    &     76.81$\pm$1.89     &    88.52$\pm$0.92    \\

\bottomrule
\end{tabular}
\end{table*}

%% file: tables/airports_results.tex
\begin{table}[htb]
\centering
\renewcommand{\arraystretch}{0.85} 

\caption{Node classification on Air Traffic Networks and BGP Network. Mean test acccuracy $\pm$ std. is shown over 20 runs.}
\label{tab:airports_bgp_results}
\begin{tabular}{@{}lccc|c@{}}
\toprule
          & Brazil & Europe & USA & BGP \; \\ \cmidrule(l){2-5}
GCN       & 64.55$\pm$4.18     & 54.83$\pm$2.69  & 56.58$\pm$1.11  & 53.33$\pm$0.18  \\
GraphSage & 70.65$\pm$5.33     & 56.29$\pm$3.21       & 50.85$\pm$2.83 & 65.19$\pm$0.28   \\
GIN       & 71.89$\pm$3.60       & 57.05$\pm$4.08       &  58.87$\pm$2.12 &  49.51$\pm$1.52 \\
Struc2vec & 70.88$\pm$4.26       & 57.94$\pm$4.01       &  61.92$\pm$2.61 & 48.40$\pm$1.39  \\ \midrule
Ours     & 76.92$\pm$5.45       & 57.12$\pm$2.81       &  63.02$\pm$1.87  & 66.54$\pm$0.48 \\ \bottomrule
\end{tabular}%
\end{table}

%% file: section_tex_files/conclusion.tex
The level of mixing plays a crucial role in characterizing real world networks. In this work we have used the quantification of local mixing patterns to study the predictive performance of graph neural networks. Our observations and results offer a new perspective to study the limitations of GNNs. Motivated by the findings of our analysis, we develop a graph transformation technique that has shown to experimentally improve assortativity owing to the use of structural regularities in the input graph and there by increasing GNN performance. Extensive experiments on various real world networks from different domains supports our claim of running GNNs on a transformed computation graph and adaptively choosing between structure and proximity information rather than on the original. The connections we find with mixing patterns and GNN learnability provides motivation for future work to provide possible theoretical claims relating the two. We also hope that this study leads to the creation of other benchmark datasets with diverse mixing patterns which can aid in the robust evaluation of future GNN methods.

%% file: section_tex_files/supplemental.tex
\section{Background}
\label{sec:nmp}
\textbf{Neural Message Passing} Concretely, during the $k^{\text{th}}$ iteration a \textit{hidden representation} $\vh_{u}^{(k-1)}$ corresponding to each node $u \in \mathcal{V}$ is updated using $u\text{'s}$ neighbourhood information. Expressed as,
\begin{equation}
\label{mpnn_eq_1}
    \vm_{\mathcal{N}(u)}^{k} = \text{AGGREGATE}^{k} \Big( \{\vh_v^{(k-1)} : v \in \mathcal{N}(u)\} \Big)
\end{equation}
\begin{equation}
\label{mpnn_eq_2}
    \vh_u^{(k)} = \text{UPDATE}^{k} \Big( \vh_u^{(k-1)}, \; \vm_{\mathcal{N}(u)}^{k} \Big)
\end{equation}
where $\text{AGGREGATE}(\cdot)$ is a trainable differentiable function mapping sets of hidden node representations of $u\text{'s}$ neighbours to an aggregated message vector, $\text{UPDATE}(\cdot)$ is also a trainable differentiable function that maps both $u\text{'s}$ current hidden representation and the aggregated message vector to $u\text{'s}$ updated representation. The initial representation $\vh_u^{(0)}$ is initialized using the original node feature $\vx_u$. After a total of $K$ iterations following Eq.~\ref{mpnn_eq_1} and \ref{mpnn_eq_2}, we obtain final representations $\vz_u = \vh_u^{K}, \forall u \in \mathcal{V} $. To perform node classification, we derive the class label of a node $u$ by decoding it's final representation via, $\evy_{u} = \text{argmax}(\text{softmax}(\text{MLP}_\theta(\vz_u)))$ where $\text{MLP}$ is a neural network with trainable parameters $\theta$ and the softmax function is used to get a probability distribution over the classes. 

\section{Applying GNN on computation graph}
\begin{algorithm}
\caption{Procedure for Node Classification}
\label{algo:our_algorithm}
    \KwIn{Computation Graph $\mathcal{C} = \big(\mathcal{V}, \mathcal{E}^\prime, \mathcal{R}, \{w_{\tau}, \forall \tau \in \mathcal{R}\}\big)$;\newline 
    Node Features $\{\vx_{u} \in \mathbb{R}^{d}, \forall u \in \mathcal{V}\}$; 
    \newline 
    1-hop $\tau$-relation specific Neighbourhood Function on $\mathcal{C}$ as $\mathcal{N}_{1}^{\tau}(u) = \{v : (u,v,\tau) \in \mathcal{E}^\prime\}$;
    }
    \KwOut{Predicted node labels}
    \Begin{
        $\vh_{u}^{0} \gets \vx_{u}\;, \forall u \in \mathcal{V} $\;
        \For{$k = 1, \dots K$}{
            \For{$u \in \mathcal{V}$}{
                
                $\vm_{u}^{k} \gets \sum\limits_{\tau \in \mathcal{R}} \sum\limits_{v \in \mathcal{N}_{1}^{\tau}(u)} \mW_{\tau}^{k} \; \vh_{v}^{k-1} \; w_{\tau}(u,v) \;  \alpha_{\tau}(u,v)$\;

                $\vh_u^{k} \gets \sigma \Big(\mW_{\text{self}}^{k} \; \vh_u^{k-1} + \; \mW_{\text{neig}}^{k} \; \vm_{u}^{k}    \Big)$

            }
            $\vh_{u}^{k} \gets \vh_{u}^{k}/ ||\vh_{u}^{k}||_{2}, \; \forall u \in \mathcal{V}$\;
        
        }
        $\vz_{u} \gets \vh_{u}^{K}, \; \forall u \in \mathcal{V}$\;
        \tcp{Predict node labels}
        \For{$u \in \mathcal{V}$}{
            $\vp_u \gets \text{softmax} (\text{MLP}_{\theta}(\vz_u))$\;
            $\evy_u \gets \text{argmax}(\vp_u)$
        }
        
        \Return{$\evy_u \;, \forall u \in \mathcal{V}$}
    }
\end{algorithm}

\section{Dataset}
\label{sec:dataset}
\input{tables/dataset_statistics}
Table~\ref{tab:dataset_statistics} provides the statistics of the datasets we use for evaluation. We select a wide range of frequently evaluated datasets from different domains and below we provide brief descriptions.

\noindent \textbf{Chameleon and Squirrel} collected by \cite{rozemberczki2019multi} are networks of hyperlinked web pages on Wikipedia related to animal topics. The nodes (here pages) are labelled from one of 5 classes based on the average traffic (views) they received. Node features are bag-of-words representation of nouns in the respective pages. We download the processed data from \citet{pei2020geom}.
    
\noindent \textbf{Actor} is a co-occurrence network based on the film-director-actor-writer network from \cite{tang2009social}. In this dataset, nodes represent actor web pages on Wikipedia and edges symbolize co-occurrence on the same web page. Node features are bag-of-words representation of the corresponding pages and labels are placed according to topics on actor web page. The dataset is from  \citet{pei2020geom}.
    
\noindent \textbf{Cornell, Texas and Wisconsin} collected as part of CMU WebKB project. Nodes are university web pages and edges are hyperlinks between them. Node labels are one of student, project, course, staff or faculty. Node features are bag-of-words representation of the corresponding web pages. The dataset is also from \citet{pei2020geom}.
    
\noindent \textbf{Cora, Citeseer and Pubmed} introduced by \cite{sen2008collective,namata2012query} are citation networks where node represent scientific papers and edges are citation relationships. Node features are bag-of-words representation of the paper and labels are the scientific field they represent.
    
\noindent \textbf{Air Traffic Networks} from three regions \textbf{Brazil, Europe and USA} is collected by the respective civil aviation agencies. Nodes represent airports and edges mean the presence of commercial routes between nodes. Nodes are labelled according to the traffic (aircraft landings and takeoffs) or level of activity (by passenger count) an airport witnesses. We get the data from \citet{ribeiro2017struc2vec}.
    
\noindent \textbf{BGP Network} collected by \cite{luckie2013relationships} represents the inter-domain structure of the Internet. Nodes represent autonomous systems and edges indicate business relationships between nodes. Node features contain categorical location and topology information and labels are based on the type or tier of the autonomous system. We get the processed data from \citet{hou2019measuring}.

\section{Practical Computation Graph Construction}
\label{algo:construction_algorithm_practical}

Algo.~\ref{algo:construction_algorithm} when naively run requires $O(n^2)$ structural similarity calculations. However in practice we use an heuristic algorithm which achieves $O(n \log n)$ calculations. The intuition is that we don’t need to look at node pairs with large degree differences. For instance, given nodes $u$ and $v$ with degree 1 and 20, we don’t have to compute the similarity between $u$ and $v$ as their structural similarity will be extremely small. We cap ourselves to a budget of $O(\log n)$ nodes to look at for each node. The budget is picked based on the heuristic of most similar degrees. The complete practical procedure is shown in Algorithm~\ref{algo:practical}.

\makeatletter 
\g@addto@macro{\@algocf@init}{\SetKwInOut{paramH}{Hyper-Params.}} 
\makeatother
\begin{algorithm}
\caption{Efficient construction of computation graph}
\label{algo:practical}
    \KwIn{Original Input Graph $\mathcal{G} = (\mathcal{V}, \mathcal{E})$; 
    \newline
    $\tau$-hop Neighborhood Function on $\mathcal{G}$ as~$\mathcal{N}_{\tau}(u):u \rightarrow 2^\mathcal{V}$; 
    Ordered Degree Sequence Function $s(V)$, $V \subset \mathcal{V}$; 
    Sequence Comparison Cost Function $\mathcal{D}$
    }
    \paramH{No. of structural relations $T < dia(\mathcal{G})$}
    \KwOut{Computation graph $\mathcal{C}$}
     \Begin{
        
        $\mathcal{E}^\prime \gets \emptyset $;$\; n \gets |\mathcal{V}|$\;
        $ w_{\tau} \in \mathbb{R}^{n \times n} \gets 0, \forall \tau \in \{0,1 \cdots T\}$;$\; w_{p} \in \mathbb{R}^{n \times n} \gets 0$\;
    
        $f_{-1} \gets 0$\;
        $S \gets [degree(u)] \; \forall u \in \mathcal{V}$ \;
        Sort S\tcp*{$O(n \log n)$}
        
        \For{$g \in \mathcal{V}$}{
            $pos \gets \text{BinarySearch} (S, degree(g))$\tcp*{$O(\log n)$}
            $P \gets \log (n)$ positions left and right of $pos$ in $S$\;
            \For{$h \in P$}{
                \tcp{P contains $O(\log n)$ nodes}
                 \For{$\tau \in \{0,1 \cdots T\}$}{
                
                    \tcp{Calculate structural dist. Eq.~\ref{eq:sd}}
                    $f_{\tau}(g,h) \gets f_{\tau-1}(g,h) + \mathcal{D} (s (\mathcal{N}_\tau(g)), s (\mathcal{N}_\tau(h)))$\;
                    
                    \tcp{Calculate edge weights.}
                    
                    $w_\tau(g,h) \gets e^{-f_\tau(g,h)}$ \tcp*{Eq.~\ref{eq:weight}}
                    
                    \tcp{Extend the edge set}
                    
                    $\mathcal{E}^\prime \gets \mathcal{E}^\prime \cup (g,h,\tau)$\;

                }
            }
        }
        \For{$(g,h) \in \mathcal{E}$}{
            $\mathcal{E}^\prime \gets \mathcal{E}^\prime \cup (g,h,p)$; $\;w_{p} (g,h) \gets 1$
            
        }
        $\mathcal{R} \gets \{0,1 \cdots T\} \cup p$\;
        \Return{$\mathcal{C} = \big(\mathcal{V}, \mathcal{E}^\prime, \mathcal{R}, \{w_{\tau}, \forall \tau \in \mathcal{R}\}\big)$}
    }

\end{algorithm}

\section{Comparison against structure aware GNNs}

\input{tables/demo-net-results}

We consider structure aware degree-based GNN models i.e. DEMO-Net \cite{wu2019net} as a baseline for node classification accuracy. DEMO-Net uses an equal number of train/val/test nodes i.e. 33\% each for the original results (Table 5 of their paper) instead of our (80/10/10) \% split (as also done in struc2vec). For fair comparison we rerun our model with the exact split and setup as demo-net. The results in Table~\ref{tab:demo_net_compare} show that our graph transformation followed by GNN approach is superior compared to structure aware GNN baseline. Although DEMO-Net utilizes degree information in message passing, the architecture is quite similar to a GIN \cite{xu2018powerful} model. In terms of expressive power, degree comparisons are usually limited to 2-3 hops as going higher leads to over-smoothing problems. Our graph transformation explicitly compares degrees at arbitrary hops and thereby provides a more efficient means to improve the assortativity of the graph.

\section{Hyper-parameter Setting}
\label{hyper_setting}
For all our experiments, we use Adam \cite{kingma2014adam} algorithm with learning rate of $\{1e-2, 1e-3\}$ and weight decay of $\{0, 1e-5, 5e-4, 5e-6 \}$ to optimize our model. Our WRGNN model variants contains 2 layers with an another fully connected MLP$_\theta$ on top of it. We select ReLU as the nonlinear activation. For the model with attention mechanism i.e. WRGAT, we use LeakyReLU with a negative input slope of 0.2. We sweep all the hidden dimensions from $\{16, 32, 64, 128\}$ for the WRGNN layer and $\{32,64,128\}$ for the final MLP$_\theta$ layer using cross validation. We set the maximum learning epochs as 500 with early stopping parameter 100. Specifically, for Hyperlinked Web Page Networks (Table~\ref{tab:main_results}) dropout operation with a probability of 0.8 is applied on each WRGNN layer. For BGP Network our model uses dropout of 0.5, learning rate as 1e-2 with weight decay of 0. Finally, for Air Traffic Networks (Table~\ref{tab:airports_bgp_results}), dropout is set to 0.6, learning rate is 1e-3 with weight decay 5e-6 and $T$ is set to 5, 5 and 8 for Brazil, Europe and USA datasets, respectively. The hyper-parameter $T$ related to structural similarity calculations is chosen based on the validation set. Its sensitivity against validation accuracy for some datasets are provided in Figures ~\ref{fig:sensitivity_structure_relations_disassortative} and \ref{fig:sensotivity_structure_relations_cora_citeseer}. 

%% file: tables/dataset_statistics.tex
\begin{table*}
\centering
\renewcommand{\arraystretch}{1.0} 
\caption{Dataset statistics. Details of $\bigstar$ discussed in Sec.~\ref{sec:exp_setup}. }
\label{tab:dataset_statistics}

\resizebox{\textwidth}{!}{%
\begin{tabular}{@{}lcccccc|ccc|ccc|c@{}}
\toprule
 & \multicolumn{6}{c|}{Hyperlinked Web Pages Network} & \multicolumn{3}{c|}{Citation Network} & \multicolumn{3}{c|}{Air Traffic Network} & \multicolumn{1}{c}{Internet Network} \\ \cmidrule(l){2-14} 
 & Chameleon & Squirrel & Actor & Cornell & Texas & Wisconsin & Cora & Citeseer & Pubmed & Brazil & Europe & USA & \multicolumn{1}{c}{BGP (small)} \\ \cmidrule(l){2-14} 
\#Nodes $|\mathcal{V}|$ & 2,277 & 5,201  & 7,600 & 183 & 183 & 251 & 2,708 & 3,327 & 19,717 & 131 & 399 & 1,190 & 10,176 \\
\#Edges $|\mathcal{E}|$ & 31,421 & 198,493 & 26,752 & 280 & 295 & 466 & 5,429 & 4,732 &  44,338 & 1,038 & 5,995 & 13,599 & 206,799 \\
\#Classes $|\mathcal{Y}|$ & 5 & 5 & 5 & 5 & 5 & 5 & 7 & 6 & 3 & 4 & 4 & 4 & 7 \\
\#Node Features $d$ & 2,325 & 2,089 & 931 & 1,703 & 1,703 & 1,703 & 1,433 & 3,703 & 500 & 1 & 1 & 1 & 287 \\
Assortativity $r_{\text{global}}$ & 0.0331 & 0.0070 & 0.0047 & -0.0706 & -0.2587 & -0.1524 & 0.7710 & 0.6713 & 0.6860 & 0.0116 & -0.0737  & 0.2629 & 0.0029 \\

Train/Val/Test Splits $\bigstar$ & \multicolumn{6}{c|}{60/20/20 } & \multicolumn{3}{c|}{60/20/20 } & \multicolumn{3}{c|}{80/10/10} & \multicolumn{1}{c}{70/10/20} \\ \bottomrule

\end{tabular}%
}
\end{table*}

%% file: tables/demo-net-results.tex
\begin{table}[htb]
\centering
\renewcommand{\arraystretch}{0.8} 
\caption{Comparison against DEMO-Net with 33 \% each train/val/test split. Node classification accuracy mean $\pm$ standard variance (same metric as DEMO-Net \cite{wu2019net})}
\label{tab:demo_net_compare}
\begin{tabular}{@{}lccc@{}}
\toprule
          & Brazil & Europe & USA \\ \cmidrule(l){2-4}

DEMO-Net (hash)       & 0.614$\pm$ 0.069& 0.479$\pm$0.064& 0.659$\pm$0.020 \\
DEMO-Net (weight) & 0.543$\pm$0.034 & 0.459$\pm$0.025 & 0.647$\pm$0.021  \\ \midrule
Ours     & 0.655$\pm$0.0055 & 0.539$\pm$0.0013 & 0.636$\pm$0.0043 \\ \bottomrule
\end{tabular}
\end{table}